\documentclass[letterpaper, 10 pt, conference]{ieeeconf}  

\IEEEoverridecommandlockouts                              

\usepackage[utf8]{inputenc} 
\usepackage[T1]{fontenc}    
\usepackage{hyperref}       
\usepackage{url}            
\usepackage{booktabs}       
\usepackage{amsfonts}       
\usepackage{nicefrac}       
\usepackage{microtype}      
\usepackage{mathtools}
\usepackage{mathrsfs}
\usepackage{multicol}
\usepackage{amsmath}
\usepackage{multirow}

\usepackage{algorithm}
\usepackage{algorithmic}

\title{\LARGE \bf
Unsupervised Domain Adaptation for Cross-Subject \\Few-Shot Neurological Symptom Detection
}

\author{Bingzhao Zhu$^{1}$ and Mahsa Shoaran$^{2}$
\thanks{*This work was supported by funding from EPFL and Cornell University.}
\thanks{$^{1}$Bingzhao Zhu is with the School of Applied and Engineering Physics,
        Cornell University, Ithaca, NY 14850, USA. E-mail: bz323@cornell.edu}%
\thanks{$^{2}$Mahsa Shoaran is with the Institute of Electrical Engineering  and Center for Neuroprosthetics,
        EPFL, Geneva 1202, Switzerland. E-mail: mahsa.shoaran@epfl.ch}
}

\begin{document}

\maketitle
\thispagestyle{empty}
\pagestyle{empty}

\begin{abstract}
Modern machine learning tools have shown promise in detecting symptoms of neurological disorders. However, current approaches typically train a unique classifier for each subject. This subject-specific training scheme requires long labeled recordings from each patient, thus failing to detect symptoms in new patients with limited recordings. This paper introduces an unsupervised domain adaptation approach based on adversarial networks to enable few-shot, cross-subject epileptic seizure detection. Using adversarial learning, features from multiple patients were encoded into a subject-invariant space and a discriminative model  was trained on subject-invariant features to make predictions. We evaluated this approach on the intracranial EEG (iEEG) recordings from 9 patients with epilepsy. Our approach enabled cross-subject seizure detection with a 9.4\% improvement in 1-shot classification accuracy compared to the conventional subject-specific scheme. 
\end{abstract}

\setlength{\textfloatsep}{0pt}

\vspace{-1mm}
\section{Introduction} \vspace{-1mm}
Machine learning (ML) has been an increasingly useful tool in neural engineering in recent years. ML can be used to analyze and classify invasive or noninvasive electrophysiological recordings, enabling timely and accurate prediction of neurological symptoms (or events) in epilepsy \cite{altaf20131, shoaran2018energy, zhu2020resot}, Parkinson's disease \cite{yao2020improved},  migraine \cite{zhu2019migraine}, and other emerging applications. 
However, despite the recent progress and potential of ML in neurological disease detection, the existing algorithms  primarily use a subject-specific scheme, requiring each patient's extended neuronal recordings to train the model. Therefore, employing such algorithms on new patients with limited labeled data has been a challenge. This is particularly the case for invasive recordings, where the duration of recording is typically short due to surgical and ethical concerns (several minutes to days).

To tackle this problem, transfer learning aims to transfer the source domain knowledge to a target domain where labeled data is difficult to acquire \cite{lotte2018review}. Over the past decade, there has been an extensive literature on domain transfer learning \cite{tzeng2017adversarial, shen2018wasserstein}, with the goal of eliminating domain shift for better generative or discriminative performance. Among transfer learning approaches, the adversarial domain adaptation introduces an adversarial loss to minimize domain shift and enforce the learned representations to share a common feature space~\cite{goodfellow2014generative}, and has obtained a promising performance in image-to-image translation tasks \cite{zhu2017unpaired,isola2017image}. Although domain adaptation techniques are widely used in computer vision tasks \cite{zhu2017unpaired,isola2017image}, their application in neural engineering and particularly in detecting neurological symptoms is still underexplored \cite{zhang2020adversarial}.

In this paper, we propose a cross-subject seizure detection algorithm based on adversarial networks \cite{goodfellow2014generative}. We mapped the features from various subjects into a subject-invariant space via the proposed unsupervised adversarial domain adaptation. Following domain adaptation, we trained an ensemble of gradient boosted trees in the subject-invariant feature space to generate cross-subject seizure predictions. The rest of this paper is organized as follows. We describe the classification task and dataset in Section II. The adversarial domain adaptation is introduced in Section III, followed by results in Section IV. Section V concludes the paper.

\vspace{-1mm}
\section{Classification Task and Data Description}\vspace{-1mm}
\label{Task}
In this work, we propose a domain adaptation model for cross-subject seizure detection. This approach was evaluated on continuous iEEG recordings from 9 patients with epilepsy. 

\vspace{-2mm}
\subsection{Seizure detection task and iEEG data}\vspace{-1mm}
Epileptic seizure detection is a supervised classification problem to differentiate between seizure and non-seizure states of a patient. We studied a total number of 97 seizure events from 9 patients. The iEEG recordings were sampled at 500Hz and annotated as \textit{seizure} or \textit{non-seizure} by domain experts (publicly available at the
IEEG Portal \cite{wagenaar2015collaborating}). All included subjects gave written informed consent and the study was approved by the
Mayo Clinic and University of Pennsylvania Institutional Review Board. We segmented the iEEG recordings of each patient to 1s windows for the subsequent processing. 
\vspace{-2mm}
\subsection{Feature extraction}\vspace{-1mm}
A set of predictive biomarkers of seizure activity \cite{shoaran2018energy} were extracted from the segmented iEEG recordings,  followed by domain adaptation and classification. The features and their definitions are as follows: line-length (LLN, $\frac{1}{d} \sum_{d}|x[n]-x[n-1]|$, $d=$ window size), total power (Pow, $\frac{1}{d} \sum_{d}x[n]^2$), variance (Var, $\frac{1}{d} \sum_{d}(x[n]-\mu)^{2},$ $ \mu=\frac{1}{d} \sum_{d}x[n]$), and band power over delta ($\delta$: 1--4 Hz), theta ($\theta$: 4--8 Hz), alpha ($\alpha$: 8--13 Hz), beta ($\beta$: 13--30 Hz), low-gamma ($\gamma_1$: 30--50 Hz), gamma ($\gamma_2$: 50--80 Hz), high-gamma ($\gamma_3$: 80--150 Hz), and ripple (R: 150--250 Hz) bands. 


\vspace{-2mm}
\subsection{Train-test split}\vspace{-1mm}
We split the data into train and test sets using a block-wise approach, in which each block is comprised of one seizure event and the subsequent non-seizure segment. To evaluate the performance, we used the first $n$ blocks for training and the remaining blocks for testing, referred to as  `$n$-shot learning'  in the following sections. The block-wise approach is a fair method to evaluate the performance, as we use a number of recorded seizure events to predict a future unseen seizure \cite{shoaran2018energy}. 

\vspace{-2mm}
\section{Adversarial Domain adaptation} \vspace{-1mm}
In this section, we consider each subject ($i$) to be associated with a specific domain ($\mathcal{D}_{i}=\left\{\mathcal{X}_{i}, P_i(\mathbf{X})\right\}$), from which features are sampled. $P_i(\mathbf{X})$ denotes the distribution of the feature vector $\mathbf{X}$. Our goal is to learn a unique encoder for each subject and map the features from this subject-specific domain to a subject-invariant domain.

\vspace{-2mm}
\subsection{Model structure}
We first consider a simple case with only two patients: one patient from the source domain ($\mathcal{D}_{S}=\left\{\mathcal{X}_{S}, P_S(\mathbf{X})\right\}$) where exists abundant labeled data for a given task, and the other patient from the target domain ($\mathcal{D}_{T}=\left\{\mathcal{X}_{T}, P_T(\mathbf{X})\right\}$) where data is expensive to acquire for the same task. 

\begin{figure}
  \centering
  \includegraphics[width=8.5cm]{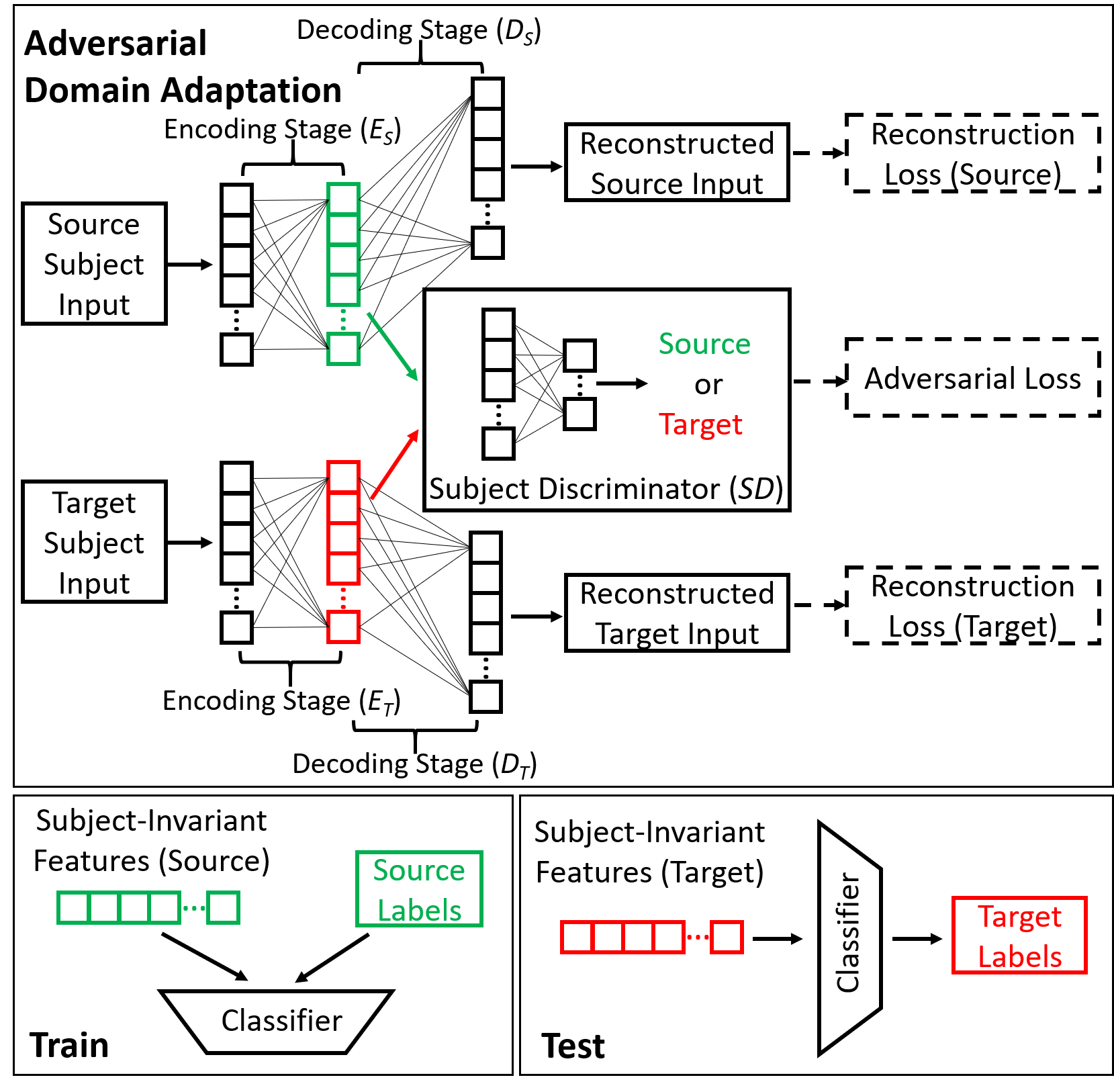}
  \vspace{-2mm}
  \caption{Model structure of the proposed unsupervised adversarial domain adaptation (top), train and test  approaches (bottom). The encoding and decoding stages form an autoencoder for each subject, which learns a latent representation of the input feature vectors. The subject discriminator takes the latent representation as input and is trained to distinguish the data from different subjects. The encoding stages are trained to fool the subject discriminator. Our goal is to learn encoders that map the input features to a subject-invariant space (denoted by green and red blocks). Following domain adaptation, a classifier is trained with the subject-invariant features to predict seizures on a target subject.}
  \label{fig1}
\end{figure}

As shown in Fig. \ref{fig1}, the proposed adversarial domain adaptation model consists of three parts: encoder (source encoder ${E}_{S}$, target encoder ${E}_{T}$), decoder (source decoder ${D}_{S}$, target decoder ${D}_{T}$), and subject discriminator (${SD}$). We used the handcrafted features ($\mathbf{X}$) as  input to the encoders. The encoders and decoders form an autoencoder, which learns a latent representation (dimension: 2048) of the original input. The subject discriminator is a multilayer perceptron, which takes the latent representations ($E_S(\mathbf{X})$ denoted by green squares, and $E_T(\mathbf{X})$ denoted by red squares) as input. The subject discriminator has two hidden layers with 512 and 128 nodes, respectively. We trained the ${SD}$ to predict whether the latent representations are from the source or target subject.

\subsubsection{Adversarial loss}
The encoders and subject discriminator form a GAN model \cite{goodfellow2014generative} for adversarial training. Here, we have encoders for both source and target domains. Let $\mathcal{L}_{\mathrm{adv}}(\mathcal{X}_{S}, \mathcal{X}_{T}, E_{S}, E_{T}, SD )$ denote the standard supervised loss of $SD$. We train the subject discriminator by minimizing the loss:

\vspace{-5mm}
\begin{gather} \label{eq1}
\min _{SD} \mathcal{L}_{\mathrm{adv}}\left(\mathcal{X}_{S}, \mathcal{X}_{T}, E_{S}, E_{T}, SD\right).
\end{gather} \vspace{-3mm}

The goal of the encoders is to minimize the distance between the empirical source and target latent representations $E_S(\mathbf{X}_S)$ and $E_T(\mathbf{X}_T)$. Thus, we trained the encoders to fool the subject discriminator and make the source/target representations indistinguishable from each other:
\begin{gather} \label{eq2}
\max _{E_{S}, E_{T}} \mathcal{L}_{\mathrm{adv}}\left(\mathcal{X}_{S}, \mathcal{X}_{T}, E_{S}, E_{T}, SD\right).
\end{gather} \vspace{-3mm}

Overall, the adversarial learning can be formalized as a maximin problem which can be solved using alternating optimization: $\max _{E_{S}, E_{T}} \min _{SD} \mathcal{L}_{\mathrm{adv}}\left(\mathcal{X}_{S}, \mathcal{X}_{T}, E_{S}, E_{T}, SD\right)$.

\subsubsection{Reconstruction loss and mode collapse}
With the adversarial training, we expect the latent space to be a subject-invariant representation of the inputs. However, the source and target encoders may simply learn to produce the same output (e.g., all zeros for latent representation), making it impossible for the subject discriminator to distinguish. In this scenario, the subject-invariant space cannot represent the inputs. This failure is referred to as mode collapse, which is a common issue with GAN training \cite{arjovsky2017wasserstein}.

To avoid mode collapse, we reconstructed the inputs from the latent representations using a decoding stage. The decoders enforce the latent representations to preserve similar information as the inputs. We calculated the reconstruction loss $\mathcal{L}_{\mathrm{rec}}$ using the mean squared error (MSE)  and $L_1$ norm was used to regularize the autoencoder.
\vspace{-3mm}
\begin{multline} \label{eq3}
\mathcal{L}_{\mathrm{rec}}(E,D)=\frac{1}{N} \sum_{n=1}^{N}\|X^{(n)}-D(E({X}^{(n)}))\|^{2} \\ + \lambda(\|E\|_{1}+\|D\|_{1}),
\end{multline} \vspace{-2mm}

\noindent where $N$ represents the mini-batch size, $X_{n}$ denotes the $n$-th sample, and $\lambda$ is the regularization coefficient, which was empirically set to 3e-5 in this work. In addition to maximizing $\mathcal{L}_{\mathrm{adv}}$, we also train the encoders and decoders to minimize the reconstruction loss. Overall, the encoders and decoders are trained with the following formula:

\vspace{-5mm}
\begin{multline} \label{eq4}
\min _{E_{S}, E_{T}, D_{S}, D{T}} -\mathcal{L}_{\mathrm{adv}}\left(\mathcal{X}_{S}, \mathcal{X}_{T}, E_{S}, E_{T}, SD\right) \\ +\alpha(\mathcal{L}_{\mathrm{rec}}(E_S,D_S)+\mathcal{L}_{\mathrm{rec}}(E_T,D_T)),
\end{multline} \vspace{-3mm}

\noindent where $\alpha$ controls the trade-off between the adversarial loss and reconstruction loss.

\vspace{-2mm}
\subsection{Multi-subject domain adaptation}\vspace{-1mm}
Previous literature on domain adaptation has only focused on transferring from one source domain to the target \cite{tzeng2017adversarial}. However, for cross-subject seizure detection, we need to consider each patient as a unique domain and transfer the feature vectors from multiple subjects to a subject-invariant domain. Here, we extended the domain adaptation framework to enable multi-subject seizure detection. The subject discriminator predicts the patient index (rather than only a single source or target), and patients are alternately considered as target while  others are considered as source. We used the cross-entropy loss for $\mathcal{L}_{\mathrm{adv}}$:
\vspace{-1mm}
\begin{gather} \label{eq5}
\mathcal{L}_{\mathrm{adv}}= - \sum_{i=1}^{N_s} \mathbb{E}_{\mathbf{X} \sim P_i(\mathbf{X})}[log(SD^i(E_i(\mathbf{X})))],
\end{gather}\vspace{-3mm}

\noindent where $N_s=9$ is the total number of patients, $SD$ outputs a vector of size $N_s$, and the $i$-th entry of the subject discriminator output ($SD^i$) indicates the probability that $\mathbf{X}$ belongs to subject $i$.

\begin{algorithm}[t]
\caption{Multi-Subject Domain Adaptation. }
\begin{algorithmic} \label{alg1}
\REQUIRE recordings from $N_s$ subjects 
\STATE $E_{1,\ldots,N_s}, D_{1,\ldots,N_s}, SD \gets$ \text{random initialization}
\FOR{number of iterations}
    \STATE Sample mini-batches of $N$ samples from all subjects $\left\{\mathbf{X}^{(1)}_1, \ldots, \mathbf{X}^{(N)}_1,\ldots, \mathbf{X}^{(N)}_{N_s}\right\}$ 
    \STATE Update the subject discriminator $SD$ by gradient decent: \\
    \vspace{-5mm}
    \begin{gather} \nonumber
    \nabla_{SD} \frac{1}{N}\sum_{i=1}^{N_s} \sum_{n=1}^{N} -log(SD^i(E_i(\mathbf{X^{(n)}_i})))
    \end{gather}
    \FOR{$i \in {1,\ldots,N_s}$}
        \STATE Update encoder and decoder $E_i$, $D_i$ by gradient decent: \\
        \vspace{-6mm}
        \begin{multline} \nonumber
        \nabla_{E_i, D_i} \frac{1}{N} \sum_{n=1}^{N} [log(SD^i(E_i(\mathbf{X^{(n)}_i}))) \\+ \alpha(\|\mathbf{X}^{(n)}_i-D_i(E_i({\mathbf{X}}_i^{(n)}))\|^{2}+\lambda(\|E\|_{i}+\|D\|_{i}))]
        \end{multline}
    \ENDFOR
\ENDFOR
\end{algorithmic}
\end{algorithm}

\subsubsection{Learning procedure}
Eq. \ref{eq1}-\ref{eq4} show the learning objectives for the domain adaptation with two subjects: a source patient and a target patient. Here, we introduce the learning procedure for multiple patients. We alternatingly considered one patient as the target and all other patients as source. The algorithmic pseudocode is shown in Algorithm. \ref{alg1}. Our goal is to leverage the labeled data from source patients to make predictions for a target patient. The domain adaptation process is essentially unsupervised, mapping different subjects' data into a common feature space. A discriminative model was trained on the subject-invariant features to generate predictions for the target patients. We tested several settings. For example, 0-shot learning does not require any labeled  recordings from the target patient. So we trained the discriminative model only on the labeled data from source patients. For $n$-shot learning, $n$ labeled seizure blocks from the target patient were used for training, in addition to the extensive data from source patients.

\subsubsection{Convergence Analysis}
The encoders map the input subject feature distribution ($P_i(\mathbf{X})$) to a latent space distribution ($Q_i(\mathbf{z})$). In the cross-subject learning scheme, we would like $Q_i(\mathbf{z})$ to be invariant across patients (i.e., $Q_i(\mathbf{z})=Q_1(\mathbf{z})$ for all $i\in {1,\ldots,N_s}$). Previous work has proven that adversarial training can reduce the shift between target and source domains \cite{goodfellow2014generative}. In this work, the subject discriminator performs a multi-class classification task and patients are alternately considered as the target (Algorithm. \ref{alg1}). Following the framework in \cite{goodfellow2014generative} 
, we recognize that Algorithm. \ref{alg1} minimizes the Jensen-Shannon Divergence ($JSD(Q_1,\ldots,Q_{N_s})$) of latent space distributions \cite{lin1991divergence}. Given that $JSD(Q_1,\ldots,Q_{N_s})$ is always non-negative and becomes zero if and only if all distributions are the same (i.e., $Q_i(\mathbf{z})=Q_1(\mathbf{z})$ for $i\in {1,\ldots,N_s}$), Algorithm. \ref{alg1} will converge to a subject-invariant space given sufficient capacity.

\vspace{-2mm}
\section{Results} \vspace{-1mm}
We tested the proposed algorithm for seizure detection from iEEG  recordings of 9 epilepsy patients. We first mapped the input feature vectors into a subject-invariant space, using the proposed unsupervised adversarial training. A discriminative model (gradient boosted trees \cite{ke2017lightgbm}) was trained in the subject-invariant space to make predictions for each patient.

\begin{figure}
  \centering
  \includegraphics[width=8cm]{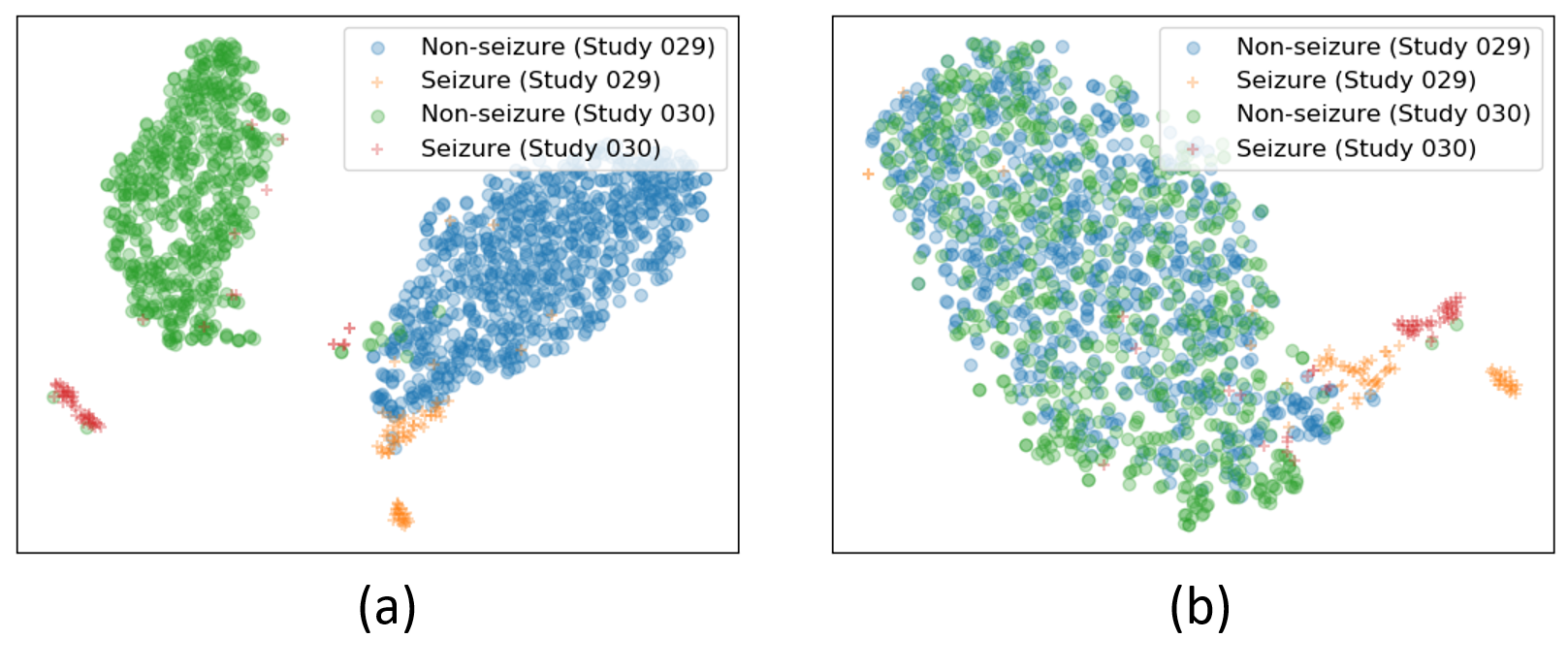}
  \vspace{-5mm}
  \caption{t-SNE visualization of the data distribution from two patients; (a) Visualization of the data distribution before domain adaptation. (b) Visualization of the subject-invariant feature space. After domain adaptation, the data from different patients become indistinguishable. }
  \label{fig2}
\end{figure}

\vspace{-2mm}
\subsection{t-SNE visualization of data distribution}\vspace{-1mm}
We used t-SNE \cite{maaten2008visualizing} to visualize the high-dimensional data distribution by mapping each data sample to a location in a 2-dimensional space. As shown in Fig. \ref{fig2}, we plot the data distribution of two patients before and after domain adaptation. We used different colors and markers to show which class the points are belonging to (seizure or non-seizure). In Fig. \ref{fig2}(a), Study 029 has a different distribution from Study 030. The domain adaptation process successfully removed the between-subject variation and brought their distributions closer to each other (Fig. \ref{fig2}(b)), enables cross-subject classification. Visualization was obtained with $\alpha=0.01$ (see $\alpha$ in Eq. \ref{eq4} or Algorithm. \ref{alg1}), which we kept for the following experiments. 

\vspace{-2mm}
\subsection{Cross-subject seizure detection}\vspace{-1mm}
We first trained the encoders to map the features into a subject-invariant space, using the domain adaptation process depicted in Algorithm. \ref{alg1}. The size of mini-batches ($N$) was set to 32. We used the Adam optimizer \cite{kingma2014adam} (learning rate of 1e-5) to update both encoders and subject discriminator for 100 epochs.
Next, 100 gradient boosted trees with a maximum depth of 4 were trained on the subject-invariant features to predict the probability of epileptic seizures \cite{ke2017lightgbm}, using the 0-shot and $n$-shot learning schemes described above. 
In the $n$-shot learning scheme, we assigned different weights to the data from source and target patients. The samples from source patients received a weight of 0.01 while the data from target patients had a sample weight of 1. We applied the reweighting scheme to address the following concerns: (1) In  few-shot learning, the training samples from source patients were more than the samples from target patients by an order of magnitude. (2) Compared to the data from source patients, the target patient data is more informative in predicting seizures on that patient.

\begin{table}[t!]\vspace{-0mm}
\caption{Performance of conventional subject-specific (SS) and cross-subject (CS) seizure detection methods.}
\vspace{-5mm}
\begin{center}
\vspace{-0mm}
\scalebox{0.77}{
\begin{tabular}{c|c|c|c|c|c|c|c}
\hline \hline
\multirow{2}{*}{Subject \#}& 0-shot & \multicolumn{2}{c|}{1-shot} & \multicolumn{2}{c|}{2-shot} & \multicolumn{2}{c}{3-shot}\\
\cline{2-8}
& CS  & SS & CS & SS & CS & SS & CS \\

\hline
\multirow{2}{*}{Study 004-2} &0.791 $\pm$  & 0.889$\pm$ & \textbf{0.935$\pm$} &\textbf{0.947$\pm$} &0.915$\pm$ &\multirow{2}{*}{N/A} &\multirow{2}{*}{N/A}  \\
&0.091 &0.047 &\textbf{0.027} & \textbf{0.033} &0.041 &&\\
\multirow{2}{*}{Study 022} &0.809 $\pm$ &0.787$\pm$ &\textbf{0.962$\pm$} &0.910$\pm$ &\textbf{0.960$\pm$} &0.875$\pm$ &\textbf{0.915$\pm$}\\
& 0.058 &0.056 & \textbf{0.012} &0.052 &\textbf{0.007} &0.050 &\textbf{0.023}\\
\multirow{2}{*}{Study 024} &0.695 $\pm$ &0.874$\pm$ &\textbf{0.949$\pm$} &0.837$\pm$ &\textbf{0.944$\pm$} &0.914$\pm$ &\textbf{0.938$\pm$}\\
& 0.124&0.030 & \textbf{0.012} &0.061 &\textbf{0.011} &0.015 &\textbf{0.012}\\
\multirow{2}{*}{Study 026} &0.715$\pm$ &0.658$\pm$  &\textbf{0.931$\pm$} &0.931$\pm$ &\textbf{0.953$\pm$} &0.928$\pm$ &\textbf{0.957$\pm$}\\
& 0.099&0.158 & \textbf{0.009} &0.011 &\textbf{0.009} &0.024 &\textbf{0.008}\\
\multirow{2}{*}{Study 029} &0.773$\pm$ &\textbf{0.813$\pm$}  &0.785$\pm$ &0.942$\pm$ &\textbf{0.944$\pm$} &\multirow{2}{*}{N/A}&\multirow{2}{*}{N/A}\\
& 0.034& \textbf{0.059} & 0.070 &0.024 &\textbf{0.022} &&\\
\multirow{2}{*}{Study 030} &0.793$\pm$ &\textbf{0.977$\pm$}  &0.974$\pm$ &\textbf{0.983$\pm$} &0.979$\pm$ &0.976$\pm$ &\textbf{0.990$\pm$}\\
& 0.044 & \textbf{0.005} & 0.010 &\textbf{0.009} &0.003 &0.025 &\textbf{0.003}\\
\multirow{2}{*}{Study 033} &0.659$\pm$ &0.888$\pm$  &\textbf{0.901$\pm$} &0.885$\pm$ &\textbf{0.887$\pm$} &0.920$\pm$ &\textbf{0.923$\pm$}\\
& 0.049 & 0.009 & \textbf{0.002} &0.005 &\textbf{0.003} &0.004 &\textbf{0.005}\\
\multirow{2}{*}{Study 037} &0.514$\pm$ &0.631$\pm$  &\textbf{0.801$\pm$} &0.597$\pm$ &\textbf{0.751$\pm$} &\textbf{0.994$\pm$} &0.989$\pm$\\
& 0.145 & 0.043 & \textbf{0.012} &0.071 &\textbf{0.037} &\textbf{0.004} &0.004\\
\multirow{2}{*}{Study 038} &0.596$\pm$ &\textbf{0.882$\pm$}  &0.858$\pm$ &\textbf{0.896$\pm$} &0.882$\pm$ &0.915$\pm$ &\textbf{0.929$\pm$}\\
& 0.029 & \textbf{0.019} & 0.027 &\textbf{0.005} &0.027 &0.011 &\textbf{0.011}\\\hline
\multirow{2}{*}{Average} &0.705$\pm$ &0.822$\pm$  &\textbf{0.899$\pm$} &0.881$\pm$ &\textbf{0.913$\pm$} &0.932$\pm$ &\textbf{0.949$\pm$}\\
& 0.030 &0.031 &\textbf{0.012} &0.012 &\textbf{0.007} &0.008 &\textbf{0.005}\\
\hline \hline
\end{tabular}
}
\label{tab3}
\end{center}
\end{table}

Given the imbalanced nature of the seizure detection task, we evaluated the classification performance using the area under the ROC curve (AUC scores). Table. \ref{tab3} compares the classification performance with/without cross-subject knowledge. In the conventional subject-specific (SS) setting, we trained the classifiers by only using the data from a target patient (e.g., $n$ seizure blocks in $n$-shot). For the cross-subject (CS) setting, we further incorporated the knowledge from source patients. We ran the proposed domain adaptation approach for 5 independent trials and reported the average performance (AUC scores) $\pm$ standard deviation. 3-shot learning on two patients (Study 004-2 and Study 029) is not applicable (N/A), since  only 3 seizure events are available in both patients. As shown in this Table, 0-shot learning achieved an average AUC score of 0.705, which is much better than the chance level (0.5). For 1-, 2-, 3-shot learning, CS outperforms the SS in terms of average classification performance. However, as we used more labeled samples from the target patient (i.e., moved from 1-shot to 3-shot learning), the difference become less significant. Overall, cross-subject learning achieved a superior performance compared to the subject-specific setting, which indicates the importance of leveraging cross-subject knowledge. In addition to seizure detection, the proposed approach has the potential to help various neurological disorders and symptom detection tasks where training data is generally limited, which remains as future work.

\vspace{-2mm}
\section{Conclusion} \vspace{-2mm}
In this paper, we proposed a novel cross-subject seizure detection framework based on adversarial domain adaptation, by mapping the features from different subjects into a subject-invariant space and applying cross-subject learning. With unsupervised domain adaptation, we achieved a better performance compared to the conventional subject-specific approach, particularly when the training data is limited (few-shot learning). The proposed model  efficiently  incorporates the knowledge from previous patients to enable high-accuracy  seizure detection in new patients. 

\vspace{-2mm}
\bibliographystyle{IEEEbib}
\bibliography{refs}

\end{document}